\documentclass[conference]{IEEEtran}
\ifCLASSINFOpdf
\else
\fi
\usepackage{subfigure}
\usepackage{hyperref}
\usepackage{booktabs}
\usepackage{amssymb,mathtools,multirow}
\usepackage{amsmath,amssymb,amsfonts}
\DeclareMathOperator*{\argmax}{argmax}
\begin{document}
%
\title{Autonomous Deep Quality Monitoring \\in Streaming Environments}

\author{\IEEEauthorblockN{Andri Ashfahani$^*$}
\IEEEauthorblockA{SCSE, NTU\\
Singapore\\
andriash001@e.ntu.edu.sg}
\and
\IEEEauthorblockN{Mahardhika Pratama$^{***}$}
\IEEEauthorblockA{SCSE, NTU\\
Singapore\\
mpratama@ntu.edu.sg}
\and
\IEEEauthorblockN{Edwin Lughofer}
\IEEEauthorblockA{DKBMS, JKU\\ Austria}
\and
\IEEEauthorblockN{E. Y. K. Yee}
\IEEEauthorblockA{SIMTech, A*STAR\\Singapore}
}


%

\maketitle

\begin{abstract}
The common practice of quality monitoring in industry relies on manual inspection well-known to be slow, error-prone and operator-dependent. This issue raises strong demand for automated real-time quality monitoring developed from data-driven approaches thus alleviating from operator dependence and adapting to various process uncertainties. Nonetheless, current approaches do not take into account the streaming nature of sensory information while relying heavily on hand-crafted features making them application-specific. This paper proposes the online quality monitoring methodology developed from recently developed deep learning algorithms for data streams, Neural Networks with Dynamically Evolved Capacity (NADINE), namely NADINE++. It features the integration of 1-D and 2-D convolutional layers to extract natural features of time-series and visual data streams captured from sensors and cameras of the injection molding machines from our own project. Real-time experiments have been conducted where the online quality monitoring task is simulated on the fly under the prequential test-then-train fashion - the prominent data stream evaluation protocol. Comparison with the state-of-the-art techniques clearly exhibits the advantage of NADINE++ with 4.68\% improvement on average for the quality monitoring task in streaming environments. To support the reproducible research initiative, codes, results of NADINE++ along with supplementary materials and injection molding dataset are made available in \url{https://github.com/ContinualAL/NADINE-IJCNN2021}.
\end{abstract}

\begin{IEEEkeywords}
evolving intelligent systems, online quality monitoring, deep learning, data streams, quality classification.
\end{IEEEkeywords}

\IEEEpeerreviewmaketitle

\section{Introduction}
\IEEEPARstart{C}{ondition} monitoring plays a vital role in today's manufacturing industries due to its positive contribution toward the increase of productivity, the safety of manufacturing operations, reduction of manpower. It maximizes the lifespan of equipment thereby avoiding unnecessary downtime leading to significant cost saving while ensuring the quality of end product preventing costly scrap, damage of workpiece or surface finishing \cite{carden2004vibration}. The worn tool is also inherent to high energy costs because it requires a high cutting force.\footnote{$^*$Equal Contribution}
\footnote{$^{**}$Corresponding Author}

Real-time quality monitoring is highly demanded because the common practice in the industry is deemed too labor-intensive as a result of multi-staged visual inspection \cite{HeidlThumfartLughoferEitzingerKlement13} \cite{RavikumaraRamachandranSugumaran11}. Accurate quality monitoring is important not only to attain high customer satisfaction but also to meet the product's standard set by the relevant authorities. Manual quality checks are limited in capacity, because of requiring time-intensive efforts in terms of personal stuff, which in turn increases costs for companies significantly. Moreover, in the complex quality control system, a manual inspection, although if conducted thoroughly and with sufficient manpower, would not cover all possible operation modes/phases with sufficient consistency and homogeneity (experts may judge the quality of items differently based on their experience levels, fatigue, mood, or even gender, etc. \cite{HeidlThumfartLughoferEitzingerKlement13}). This issue has led to an in-depth study in utilizing the data-driven approaches to fully automate quality monitoring of manufacturing products, which abandons the usage of manual checks and induces human-like inconsistencies.  

Data-driven quality monitoring is developed through two steps \cite{Mitra16}, indirect sensing and monitoring where the indirect sensing phase is designated to extract notable features of mounted sensors while the monitoring phase utilizes extracted features to train a model autonomously from the data, which circumvents high development time as is the case for analytical, physical-oriented models. 
The models are capable of diagnosing possible defects of end products \cite{Montgomery08} without laborious manual intervention. 

Various approaches have been presented in the literature to deliver a reliable data-driven quality monitoring approach. In \cite{paul2012multi}, a neural network is developed to predict the cutter's flank wear in the metal-turning process. A paper investigates the application of an acoustic emission sensor to produce informative features for condition monitoring of chemical-mechanical planarization (CMP) \cite{jeong2006multi}. In \cite{li2009fuzzy}, the vibration sensor is made use in predicting the tool condition of the ball-nose end-milling process and combined with the fuzzy regression technique. A quality monitoring approach in the multi-staged manufacturing process is proposed in \cite{khormali2016novel}, using the clustering approach and the support vector machine. The techniques proposed in \cite{SerdioLughoferPichlerBucheggerEfendicJour13,SerdioLughoferPichlerPichlerBucheggerEfendicJour15} focus on the detection and localization of faults in rolling mills by relying on an all-coverage data-driven modeling approach to integrate as many sensor signals as possible in a whole network of models. 

Since the multi-sensors are implemented and often lead to the curse of dimensionality, feature selection is needed \cite{StanczykJain16} \cite{GuyonElisseeff03} to reduce the dimension of feature space. Despite its rapid progress, all of these approaches are deemed as offline approaches and assume stationary environments and operation modes. A model is fixed once trained thereby resulting in major performance deterioration in the presence of concept drift \cite{PratamaLuLughoferZhangEr16}, an occurrence that leads to changing data distributions and even changing input-output (target) relationships (thus, older trained relations become outdated), see \cite{KhamassiMouchawehHammamiGhedira16} for a survey. 
Consequently, a retraining phase has to be regularly carried out by experts/operators, which incurs extra costs and intensive computational resources.   

The data-driven condition monitoring approach has been advanced in \cite{PratamaDimlaLughoferPedrycz18} via the deployment of data stream algorithm, pENsemble+ featuring the self-evolving structure and one-pass learning principle. Moreover, pENsemble+ is equipped with the online active learning strategy and the online feature selection approach using ensemble classifiers in an evolving data-streaming context \cite{PratamaPedryczLughofer18} to permanently update their performance and to handle with non-stationary modes. Another online quality monitoring methodology is proposed in \cite{LughoferZavoianuPollakMeyerHeyeEitzingerRadauer18jour} for microfluidic chip quality. It integrates the incremental partial least square (iPLS) method into GEN-SMART-EFS \cite{LughoferCernudaKindermannPratama14} for online dimensionality reduction strategy of high-dimensional multi-sensor data. This approach is extended in \cite{LughoferZavoianuPollakPratamaHeyeZoerrerEitzingerRadauer18} by introducing the forgetting strategy to handle the concept drift and the multi-objective evolutionary computation for process optimization. A collection of further dynamic data-driven QC approaches can be found in \cite{LughoferMouchaweh19}, which all serve as one key aspect in nowadays predictive maintenance systems.
The online quality monitoring topic deserves further in-depth study due to at least two rationales: 
\begin{enumerate}
\item Existing approaches heavily rely on the tedious feature engineering step of sensory data. This leads the overall framework to be application dependent with significant development time needed and high input dimension calling for the feature reduction mechanism. Such approaches cannot operate in an end-to-end manner. Notwithstanding that the application of deep learning starts to emerge in the literature where they resolve the issue of hand-crafted features and curse of dimensionality due to its implicit feature learning trait via multiple nonlinear transformation \cite{goodfellow2016deep}, approaches employing deep learning hardly scale to the streaming environments and become outdated quickly in the rapidly changing environments; 
\item Existing approaches do not take into account the heterogeneous source of information leaving aside the use of multi-modal information, e.g., sensor+image, sensor+text. In order to assure accurate diagnosis of product's defect, especially in systems where the prediction quality depends on the mixture of various process phases or typically records different types of data, a multi-modal model can be utilized \cite{guo2017hybrid,zhao2019deep}.
\end{enumerate}

This study presents an online quality monitoring of transparent mold within a non-stationary streaming environment using an extension of a recently developed deep learning algorithm for data streams, termed as {\em Neural Network with Dynamically Evolved Capacity++} (NADINE++). The modification of NADINE in \cite{nadine} encompasses the integration of 1-D and 2-D convolutional layers (CNN) to bypass a complex feature engineering step in the condition monitoring task. The 2-D CNN is taken from the pre-trained residual network (ResNet) in the ImageNet problem. That is, only those generating general features are adopted leaving aside the object-specific layers. NADINE++ itself features an autonomous trait where the fully connected layer is self-evolved from scratch with the absence of a predefined network structure. In the classifier part, both the hidden nodes and hidden layers are automatically evolved or removed overtime via the network significance (NS) method and the drift detection method, respectively. The concept of adaptive memory exists to perform the experience replay mechanism combating performance loss during the insertion of a new layer. In addition, the soft forgetting mechanism is prepared to overcome the exploding gradient problem by controlling the step size of model updates. 

The online quality monitoring task is performed in two facets: the use of sensory information from the sensor and the use of multi-modal information from both sensory and visual information. The first one is handled by amalgamating 1-D CNN while the second one is addressed by both 1-D and 2-D CNNs. This is established by creating a heterogeneous network in which the output of 2-D CNN is fused with the sensory data processed by 1-D CNN and in turn passed to the evolving multilayer perceptron (MLP) classifier. Furthermore, a multi-class online quality classification problem is studied here realizing 2 implementation scenarios: one-step-ahead and current-batch prediction. The explanation about the injection molding machine can be found in the supplemental document.

The major contribution of this paper is summed up into three folds: 1) this paper offers NADINE++ for online quality monitoring in truly streaming environments. NADINE++ is free from complex feature engineering step and enables the end-to-end training mechanism while being self-adaptive to track changing distributions of data streams; 2) this paper offers a comprehensive study for online quality monitoring of transparent mold from the injection molding machine where it investigates the use of sensory information, image information and multi-modal information. In addition, multi-class online quality classification schemes can be handled, providing a direct fault identification step (through the prediction of fault classes); 3) real-time experiments on the injection molding machine have been carried out where all real-world data, Python implementation of NADINE++, supporting materials are made publicly available for the convenience of reproducing our results. Comparisons with several state-of-the-art algorithms exhibit the advantage of NADINE++ in all simulation scenarios.

\section{Problem Formulation}
Online quality monitoring is defined here as the detection problem of transparent mold quality produced by the injection molding machine. Unlike in the offline case where a predictive model $f(.)$ is crafted based on prerecorded samples $D=[X;Y]\in\Re^{(N\times (u+m))}$, where $N,u,m$ respectively stand for the number of data points, input features and target classes, and fixed once deployed, the online quality classifier is changed with streaming data. That is, there exist continuous information flows $B_1,B_2,\dots,B_k\dots,B_K\in\Re^{N\times u}$ where $K$ denotes the number of batches. The typical characteristic of the data stream is seen in the rapidly changing data distributions, i.e. $P(X,Y)_k\neq P(X,Y)_{k+1}$ \cite{GamaDataStream}. Adapting to such changes without a retraining phase from scratch may catastrophically erase previously valid knowledge as only forces to be trained to the new one. In addition, the retraining phase imposes considerable computational and memory footprints that cannot keep pace with the fast characteristics of the molding machine. A model is often forced to predict the data batch $B_k$ first due to the issue of label latency. The prequential test-then-train protocol is implemented in this study where a model $f(.)$ is used to predict the target of the data batch $B_k$ before updating it with the same data batch \cite{GamaDataStream}. Numerical evaluation is executed per data batch $B_k$ in order to check the effect of concept drift in the numerical results.

Input attributes $X\in\Re^{N\times u}$ arrive with the absence of target classes where it is collected from the indirect sensing mechanism \cite{cazalas2016modulation} via built-in sensors of the injection molding machine generating time-series data and external static camera mounted at the end of manufacturing cycle delivering colored photos of the transparent mold. The data batch $B_k=X_k$ are collected in batch within the sampling time leading to $N=50$ and $K=59$. Predictions are made for this current batch of samples to produce classification statements (labels with uncertainties) based on which an expert/operator can provide feedback. These in turn can be used as annotation labels collected in a target vector $Y_k$, resulting in a supervised batch of samples $B_k=[X_k;Y_k]$. The labeled data batch $B_k$ is then used for training the deep network in order to be up-to-date with the most recent batch for reliably classifying the next batch. Our implementation comprises two scenarios, one-step-ahead and current-batch prediction. Both are multi-class classification problems where it consists of three classes: good, short-forming and weaving, leading to $m=3$ as explained in the supplemental document. 

\section{Learning Policy of NADINE++}
\subsection{Network Architecture of NADINE++}
NADINE++ extends the original NADINE in \cite{nadine} with the addition of the feature extraction layer alongside the self-evolving fully connected layer generalizing its feasibility into various condition monitoring problems regardless of the machine or sensor specification. That is, the raw input samples $X$ collected from built-in sensors or external cameras are fed to the convolutional feature mapping $F(X)$ extracting the feature mapping. The residual mapping \cite{resnet18} is implemented here where the underlying goal is to approximate a residual function $H(X)-X$ where $H(X)$ is the desired mapping function. This concept is similar to the introduction of a shortcut connection in terms of the identity mapping to the convolutional feature $F(X)+X$. The use of such a connection is inspired by the increase of difficulty in training a very deep network under a standard structure due to the issue of vanishing or exploding gradient problem. It is also supported by the finding in \cite{zhang2018information} where the mutual information of the natural features rapidly decreases across the network depth. The output of $l-th$ residual building block $Z\in\Re^{u'}$ is formalized as follows:
\begin{equation}
    Z_l=F(X,W_{conv}^{l,i})+X
\end{equation}
where $u'$ denotes the number of natural features, while $W_{conv}^{l,i}$ stands for the $i-th$ filter weight of the $l-th$ layer. Note that two variants of filters is deployed here, termed one dimensional $W_{conv}^{l,i}\in\Re^{g}$ filter and two dimensional  $W_{conv}^{l,i}\in\Re^{g\times g}$ filter where $g$ denotes the filter size. It is put forward to construct the 1-D and 2-D CNN network structure handling time-series and visual information, respectively \cite{Du_2020}.

After stacking $L_{conv}$ residual layers, the natural features of the last residual layer is passed to the classification layer formed as a MLP network parameterized by the connective weight and bias $W_{in}^{i,l}\in\Re^{u'\times R_l},b_{i,l}\in\Re^{R_l}$ where $R_l$ stands for the number of hidden units in the $l-th$ fully connected layer. It is formally written as follows:
\begin{equation}
    h_{l}=s(W_{in}^{i,l}Z_{L_{conv}}+b_l)
\end{equation}
Note that the fully connected layer of NADINE++ features a self-adaptive trait allowing the number of hidden nodes $R_l$ and the number of fully connected layers $L_{fully}$ to be automatically developed during the training process. The classification decision is drawn by a softmax layer converting the activation degrees of the last hidden layer into the output posterior probability as follows:
\begin{equation}
    \hat{y} = softmax{(W_{out} h_{L_{fully}} + c)}
\end{equation}
where $W_{out}\in\Re^{R_{L_{fully}} \times m}, c\in\Re^{m}$ are the output weights and the output bias. The predicted label $\hat{Y}$ is taken from the output $\hat{y}$ having the highest multi-class probability. This can be written mathematically as $\hat{Y} = \argmax_{\hat{y} \in \Re^m} \hat{y}$.

\subsection{Structural Learning of NADINE++} NADINE++ is constructed by an evolving MLP classifier. It features an open structure where its hidden nodes and layers are automatically generated based on the network significance (NS) method and the drift detection method pinpointing possible changes in data distributions. The NS method is crafted using the bias-variance decomposition signifying the underfitting and overfitting situations. A new node is introduced in the case of underfitting or high bias while an inconsequential node is pruned if the overfitting or high variance condition is present. Network bias and variance are expressed as follows:
\begin{eqnarray}\label{NS}
   NS = E[(\hat{y}-E[\hat{y}]^2)]+(E[\hat{y}]-y)^2\\
   NS = Variance(\hat{y})+Bias(\hat{y})^2 
\end{eqnarray}
where $Variance(\hat{y})$ and $Bias(\hat{y})$ respectively stand for the network variance and bias respectively. Equation (\ref{NS}) is solved by assuming that $x$ is drawn from the normal distribution $p(x)=\frac{1}{\sqrt{2\pi\sigma}}\exp{(-\frac{(x-\mu)^{2}}{\sigma^{2}})}$ with the mean $\mu$ and variance $\sigma$. Unlike in the original NADINE, $x$ here denotes the features generated from the convolutional layer.

The statistical process control (SPC) condition \cite{macgregor1995statistical} is extended by integrating the adaptive confidence level controlling the degree of confidence with respect to the level of network bias and variance. It is written as follows:
\begin{eqnarray}
    Growing: \mu_{bias}^{t}+\sigma_{bias}^{t}\geq \mu_{bias}^{min}+\kappa\sigma_{bias}^{min}\label{Growing}\\
    \kappa=1.25\exp{(-Bias^{2})}+0.75\\
    Pruning: \mu_{var}^{t}+\sigma_{var}^{t}\geq \mu_{var}^{min}+2\xi\sigma_{var}^{min}
    \label{Pruning}\\
    \xi=1.25\exp{(-Var^{2})}+0.75
\end{eqnarray}
It is worth noting that the term 2 is inserted into (\ref{Pruning}) to avoid the direct-pruning-after-adding condition. This strategy enables dynamic confidence level in the range of $[68.2\%,95.2\%]$ and $[68.2\%,99.9\%]$. Furthermore, $\mu_{bias}^{min},\sigma_{bias}^{min},\mu_{var}^{min},\sigma_{var}^{min}$ are reset once (\ref{Growing}) and (\ref{Pruning}) are satisfied. It allows the node growing and pruning mechanisms to be carried out frequently in the case of high bias and variance. The two conditions are capable of capturing abnormal patterns leading to the increase of network bias and variance.  It is worth mentioning that the node growing and pruning phase is only localized in the last layer to enable stable hidden representation. 

The hidden layer growing strategy of NADINE++ is controlled by the drift detection approach based on the Hoeffding's bound strategy to control the appropriate level to declare a drift. It is modified from \cite{PratamaDimlaLughoferPedrycz18} where it utilizes the accuracy vector $A_k\in\Re^{n}$ recording the prequential error of the $k-th$ data batch $B_k$. Misclassified instance is marked as "1" while correct classification returns "0". As a result, the increase of population mean mirrors performance's deterioration of a model thereby signifying the concept drift. The first step is to determine the switching point $cut$ signalling the start of concept drift $cut$:
\begin{eqnarray}
\hat{A}+\epsilon_{\hat{A}}\leq\hat{B}+\epsilon_{\hat{B}}
\end{eqnarray}
where $A\in\Re^{N}$ and $B\in\Re^{cut'}$ and $cut'\leq N$. $cut'$ stands for the hypothetical cutting point defined as $[25\%, 50\%, 75\%]*N$ to address the issue of false alarm while $\hat{A}$ and $\hat{B}$ denote the statistics of accuracy matrix $A$ and $B$, respectively. The Hoeffding's bounds $\epsilon_{\hat{A}},\epsilon_{\hat{B}}$ are derived as follows \cite{frias2014online}:
\begin{equation}
    \epsilon_{\hat{A},\hat{B}}=\sqrt{(1/(2n_{\hat{A},\hat{B}}))\ln (1/\alpha)}
\end{equation}
where $n_{\hat{A},\hat{B}}$ denotes either number of sample in $A$ or $B$; and $\alpha$ is the significance level of the Hoeffding's bound and is inversely proportional to the confidence level \cite{frias2014online}. 

Once eliciting the cutting point $cut$, it enables the construction of another partition of accuracy matrix $C\in\Re^{(N-cut)}$, where $A=[B;C]$. A drift is signalled if $|\hat{B}-\hat{C}|\geq \epsilon_{d}$ holds in which it compares two possible distinct concepts carried in $B$ and $C$. Another situation is formulated in the warning condition likely leading to the drift case but still calling for confirmation with next data streams. It is set as $|\hat{B}-\hat{C}|\geq \epsilon_{w}$. The Hoeffding's bounds to confirm warning and drift condition is derived as follows \cite{frias2014online}:
\begin{equation}
    \epsilon_{d,w}=\sqrt{\frac{N-cut}{2\times cut\times N}ln(\frac{1}{\alpha_{d,w}})}
\end{equation}
Since $\epsilon_{d}>\epsilon_{w}$, the significance level is set as $\alpha_{d}<\alpha_{w}$ \cite{frias2014online}. On the other hand, the stable condition is returned if the null hypothesis remains valid. 

A new layer is inserted in the case of drift thereby deepening the network structure while a data buffer $B_{w}=[B_{w};B_k]$ is created in the warning phase. The data buffer is to be replayed if the drift case is substantiated in the next data batch checking the consistency of the data buffer. Only the parameter learning phase is carried out during the stable phase. It is worth mentioning that the addition of a new layer must be undertaken carefully due to the risk of catastrophic forgetting. That is, an untrained layer is stacked to the last layer playing a vital role in the network's output. The adaptive memory concept is applied to overcome this issue inspired by the experience replay mechanism of continual learning \cite{nadine}. 

\subsection{Adaptive Memory Strategy}
The addition of a new layer in the classifier induces the catastrophic forgetting problem since the final output is controlled by an untrained component missing previous concepts. The adaptive memory strategy is implemented here where it performs the experience replay-like mechanism \cite{schaul2015prioritized}. The key idea is to memorize important samples describing the underlying data distribution. Important samples are defined as those satisfying one of the two conditions: 1) they must not be redundant samples leading to the over-fitting issue; 2) they must not be outliers portraying the low variance direction of data distribution. The concept of ellipsoidal anomaly detector is applied here \cite{moshtaghi2009anomaly} where it enables to select of anomalous but useful samples based on the multivariate Gaussian distribution. A sample is selected into the adaptive memory if the following condition is satisfied: 
\begin{equation}\label{ellipsoidal}
    t^{1}_{u}\leq M(x_k;C,Cov^{-1})\leq t^{2}_{u}
\end{equation}
where $C$ and $Cov^{-1}$ respectively stand for the center and inverse covariance matrix of the multivariate normal distribution, while $t^{1}_{u}$ and $t^{2}_{u}$ denote the inverse of the chi-square distribution with $u$ degrees of freedom $\chi^{2}_{u}(\alpha)$; $M$ denotes the Mahalanobis distance. $\alpha$ is the confidence level and set such that $\alpha_1<\alpha_2$ where $\alpha_1=0.99$, $\alpha_2=0.999$. This means that edge points of the data distribution are selected characterizing well the outer contour of the distribution, but no real outlying points. Note that the threshold selection from any unimodal distribution encompasses the majority of data points.

Since the main goal of (\ref{ellipsoidal}) is to capture unique samples, it often leads to too few samples being captured thus making the experience-replay mechanism ineffective. In addition, hard samples are also picked up when they characterize low confident samples lying close to the decision boundary, i.e. $P(Y|X)\approx 0.5$.
Such samples are important for the training procedure in order to 'sharpen' the decision boundary appropriately and thus to reduce the likelihood of false classifications. 
A sample is stored in the adaptive memory if the following condition holds:
\begin{equation}
    \frac{\hat{y}_1}{\hat{y}_1+\hat{y}_2}\leq \delta
\end{equation}
where $y_1,y_2$ denote the highest and second highest predictive output while $\delta$ stands for the predefined threshold fixed at 0.55. Uncertain output is resulted from low ratio between $y_1$ and $y_2$ or is close to $0.5$.

\subsection{Soft Forgetting Strategy}
The soft forgetting mechanism is put forward to govern the learning intensity of MLP classifier components via the adjustment of learning rate. It copes with the catastrophic forgetting problem affecting the performance of NADINE++ and the exploding gradient problem leading to saturating hidden nodes. Intuitively, all relevant parameters to a batch $B_k$ should be updated. On the other hand, the amount of update to all irrelevant parameters should be minimized \cite{yoon2017lifelong,rusu2016progressive}. This is realized by measuring the correlation of every node of the $l-th$ layer to the target variable on every incoming batch $B_k$ in which the current concept is embraced. It is followed by setting the highly correlated layer's learning rate to a high value while hindering other layers to accept the current concept. The learning rate of the $l-th$ layer $\eta_{l}$ is formulated as follows:
\begin{equation}
\eta_{l}=0.02*\exp{(-(1/\rho(h_{l},\hat{y})-1))}    
\end{equation}
where $\rho(h_{l},Y)$ denotes the average correlation between all nodes of the $l-th$ layer to all outputs calculated by the Pearson correlation index. In this study, $\eta_{l}$ was capped at $0.02$. 

\section{Results}
This section demonstrates the classification performance of NADINE++ on injection molding data. A prequential test-then-train procedure \cite{GamaDataStream} was used to demonstrate NADINE++'s performance where each data batch is tested first before it is learned. The performance of NADINE++ was simulated on 2 scenarios: One-step-ahead and current-batch prediction. The former scenario tests NADINE++'s ability to estimate the next batch product quality $Y_{k+1}$ exploiting the current batch sensor data $X_{k}$. The latter scenario is the main proof of concept attempting to validate NADINE performance in predicting the current batch mold condition $Y_{k}$ based on current sensor and image data $X_{k}$. Also, an ablation study is presented to assess NADINE++'s components. 

\subsection{One-step-ahead Prediction}
\subsubsection{Network Structure} 
\begin{figure}[!t]
\centerline{\includegraphics[scale=0.4]{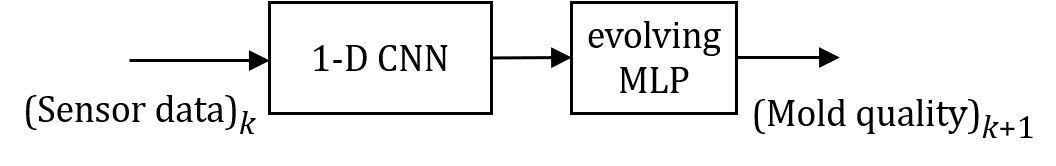}}
\caption{The NADINE++ architecture used in the first scenario.}
\label{fig:scenario1}
\end{figure}

The proposed NADINE++ network structure to handle the one-step-ahead prediction problem consists of 2 main components, i.e., 1-D CNN as a feature extractor and an evolving MLP as a classifier, as illustrated in Fig. \ref{fig:scenario1}. 1-D CNN was constructed by stacking 3 convolutional layers consisting of 1-D kernels with a pixel, padding and stride. The number of input and output channels in the first, second and third layers were respectively selected as $[48,60]$, $[60,40]$ and $[40,20]$. The learned features were then forwarded to an evolving MLP which is able to construct its structure based on problem complexity. A ReLU non-linear activation function was used to decouple the layers. Note that in this scenario, NADINE++ predicts the next batch mold quality $Y_{k+1}$ based on the current batch sensor data $X_{k}$.

\subsubsection{Algorithms and Parameters}
The proposed NADINE++ model is compared with other classification algorithm such as Online Deep Learning (ODL) \cite{sahoo2017online}, NADINE \cite{nadine} and incremental bagging \cite{IncBaggingBoosting}. Those are online learning algorithms that use a single scan to process the incoming data. ODL is built upon MLP having a different-depth structure. It is also empowered by hedge backpropagation which is able to improve the performance of the deep networks. The comparison against NADINE aims to present the improvement over an evolving MLP classifier without a feature extractor. The performance of NADINE++ is also benchmarked against the performance of Incremental Bagging.

Those algorithms are re-implemented in the same simulation scenario and procedure to ensure a fair comparison. In the implementation, NADINE, ODL and NADINE++ are trained by Stochastic Gradient Descent (SGD). The hyperparameters are re-tuned for each algorithm attempting to obtain better performance, thereby providing a more competitive experimental setting for testing out our proposed algorithm. NADINE++ parameters $\alpha_{d}$ and $\alpha_{w}$ are set to $0.0001$ and $0.0005$ which controls the drift rate. The learning rate and momentum coefficient are set as $[0.001,0.02]$ and $0.95$. All of these parameters remain unchanged in all experiments. The documentation can be checked in the aforementioned link.

\subsubsection{Results} Table \ref{result-onestepahead} shows the performance of consolidated algorithms.
The performance is evaluated according to two criteria: accuracy and number of layers (depth). These metrics enable the user to observe the predictive performance of an algorithm and its complexity. The tests were performed 10 times. The average performance across 10 consecutive runs is listed in the table. The t-test is conducted to further validate NADINE++'s predictive performance. The $\times$ mark in the table indicates that the t-test rejects the null hypothesis at the 5\% significance level. These are also applied in the second case. 

From Table \ref{result-onestepahead}, it can be observed that NADINE++ outperforms NADINE, ODL and Incremental Bagging in terms of accuracy. Further, the t-test confirms that this result is statistically significant (P $<$ 0.05). This is understood as NADINE++ employs 1-D CNN as a feature extractor. Note that the sensor data used in this experiment is the original normalized time-series signal. There is no signal processing method applied to the sensor data before 1-D CNN. NADINE++ surpasses other methods, indicating the advantage of 1-D CNN when processing sensor data. By comparing NADINE++ and Incremental Bagging, we see the benefit of deep neural network architecture to execute a classification task in a streaming environment.

In terms of network complexity, NADINE++ introduced less number of layers compared to NADINE. It is worth mentioning that these algorithms utilize the same drift detection mechanism to govern the addition of depth. This finding signifies that 1-D CNN is able to produce more robust features for the classifier \cite{wang2015transferring}. These features help NADINE++ to minimize performance deterioration due to concept drift. As a result, there is less drift detected by NADINE++ which in turn induces less number of the hidden layer. In industrial applications, it is important to maintain the network complexity low to attain high feasibility for real-time deployment \cite{he2018amc}.
\begin{table}
\caption{Classification performance on One-step-ahead Prediction Scenario}
\label{result-onestepahead}
\begin{center}
\scalebox{1}{
\begin{sc}
\begin{tabular}{lccr}
\toprule
Model & Depth &  Acc. (\%) \\
\midrule
NADINE    & 9.2 $\pm$ 1.69 &  76.46 $\pm$ 0.03$^\times$ \\
ODL & 5 & 79.40 $\pm$ 0.02$^\times$ \\
Incremental Bagging & N/A & 81.65 $\pm$ 0.00$^\times$ \\
NADINE++ & 6.4 $\pm$ 0.70 & \textbf{83.40 $\pm$ 0.02} \\
\bottomrule
\end{tabular}
\end{sc}}
\end{center}
\centering{}\footnotesize{$^{\times}$: Indicates that the numerical results of the respected baseline and NADINE++ are significantly different.}
\end{table}

The precision and recall of each class are evaluated. From Table \ref{result-onestepahead-precrec}, it is depicted that there are gaps between precision and recall. However, the precision and recall values of NADINE++, which are never less than 0.75, can be considered good for a 3-class classification problem. Also, the gaps are quite small indicating that the predictive performance of NADINE++ is unbiased to one of the classes. One may re-tune the hyperparameters to improve the precision and recall metrics in one class. However, it should be conducted carefully as there is a trade-off between precision and recall \cite{virtanen2019precision}.
\begin{table}
\caption{Precision and Recall of NADINE++ on One-step-ahead Prediction Scenario}
\label{result-onestepahead-precrec}
\begin{center}
\scalebox{1}{
\begin{sc}
\begin{tabular}{lccr}
\toprule
Labels & Precision & Recall \\
\midrule
Normal & 0.84 $\pm$ 0.044 & 0.89  $\pm$ 0.032 \\
Weaving & 0.83 $\pm$ 0.019 & 0.77 $\pm$ 0.071 \\
Short-forming & 0.84 $\pm$ 0.031 & 0.85 $\pm$ 0.063 \\
\bottomrule
\end{tabular}
\end{sc}}
\end{center}
\centering{}
\end{table}

\subsection{Current-batch Prediction}
\subsubsection{Network Structure} 
\begin{figure}[!t]
\centerline{\includegraphics[scale=0.4]{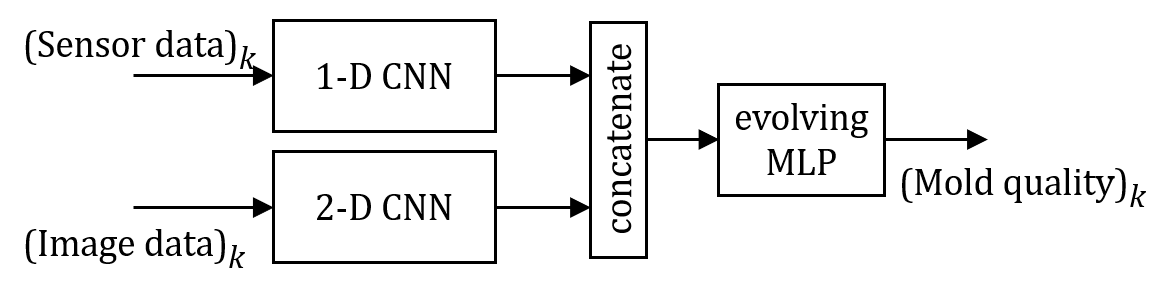}}
\caption{The NADINE++ architecture used in the second scenario.}
\label{fig:scenario2}
\end{figure}

The second scenario requires NADINE++ to predict the mold quality $Y_{k}$ based on the current batch sensor and image data $X_{k}$. In the production system, this mechanism is useful as the technician can monitor the product quality without interrupting the molding process. As depicted in Fig. \ref{fig:scenario2}, NADINE++ put forwards three main components to handle this classification problem, those are 1-D CNN, 2-D CNN and an evolving MLP classifier. The first and second component function to extract useful features from sensor and image data, respectively. The same 1-D CNN structure as used in the first scenario was utilized here, whereas the 2-D CNN structure adopted the ResNet18 feature extractor part. The simplest version of ResNet was selected attempting to reduce the risk of overfitting. ReLU activation function was used to introduce non-linearity. Finally, the generated features by 1-D and 2-D CNN were concatenated together into a long vector for the input of the evolving MLP classifier.

\subsubsection{Algorithms and Parameters} We compared NADINE++ with four deep network architectures: NADINE \cite{nadine}, ODL \cite{sahoo2017online}, ResNet18 \cite{resnet18} and VGG11 \cite{vgg11}. The first two methods were designed to handle the classification problems in an online manner. These methods are able to incrementally improve predictive performance as the number of data increases. NADINE and ODL do not implement any feature extraction layer. As a result, they can only exploit raw sensor data in this scenario with the absence of automatic feature engineering trait. Note that the generated image data from our experiment are RGB images with a size of $150 \times 150 \times 3$. One may easily convert it to a vector with a size of $67500 \times 1$ to fit in NADINE and ODL. In practical implementation, however, a huge number of inputs may decrease the performance of NADINE and ODL due to the over-fitting problem \cite{liu2005toward}. 

ResNet18 and VGG11 are the prominent deep network architecture for computer vision. Both of them can achieve more than 80\% accuracy on ImageNet with  \cite{belilovsky2019greedy}. In this scenario, these networks only used image data to predict the mold quality as both of them did not incorporate 1-D CNN. We selected the simplest structure of ResNet and VGG because the number of collected data in our scenario is quite low. This may lead to overfitting if more complex networks were used \cite{baum1989size}. All baselines were re-implemented in the same simulation scenario to provide an equal comparison. We used the SGD method to performs end-to-end training on all algorithms. For the baselines, their hyperparameters are hand-tuned and the best-performing results are reported here. For NADINE++, we used the same hyperparameters as used in the first scenario. These settings can also be observed in the raw numerical results which have been uploaded in the provided link.

\subsubsection{Results} Table \ref{result-currentbatch} shows the predictive performance of the second scenario. It is observed that NADINE++ achieves the best performance in terms of accuracy. From the t-test, we also found that NADINE++'s performance was statistically different (P $<$ 0.05) compared to NADINE, ODL and VGG11. This is reasonable as NADINE++ is able to process both sensory and visual information \cite{Du_2020}. Also, the end-to-end training mechanism enables NADINE++ to put more attention on which features help to improve the performance. The performance of NADINE++ is not statistically different from ResNet18. This is understood as NADINE++'s 2-D CNN adopted from ResNet18. However, it is presented in Table \ref{result-currentbatch-precrec} that our method achieved better precision and recall compared to ResNet18. In addition, NADINE++ is able to increase its capacity if the problem getting more complex.
\begin{table}
\caption{Classification performance on Current-batch Prediction Scenario}
\label{result-currentbatch}
\begin{center}
\scalebox{1}{
\begin{sc}
\begin{tabular}{lccr}
\toprule
Model & Depth &  Acc. (\%) \\
\midrule
NADINE$^*$ & 8.9 $\pm$ 1.45 & 76.9 $\pm$ 2.28$^\times$ \\
ODL$^*$ & 5 & 80.4 $\pm$ 0.01$^\times$ \\
\midrule
ResNet18$^{\#}$ & 17 &  86.5 $\pm$ 0.85 \\
VGG11$^{\#}$ & 11 &  84.1 $\pm$ 0.99$^\times$ \\
\midrule
NADINE++ & 20 $\pm$ 0.47 & \textbf{87.1 $\pm$ 0.74}\\
\bottomrule
\end{tabular}
\end{sc}}
\end{center}
\centering{}\footnotesize{$^*$: Exploits sensor data, $^{\#}$: Exploits image data.}
\end{table}

Compared to NADINE, our method generated less number of layers. NADINE++ added around 3 hidden layers, whereas NADINE introduced more than 8 layers. Note that the reported number of layers in Table \ref{result-currentbatch-precrec} is the total number of layers including the 2-D CNN layers. Interestingly, the same approach was utilized to control the hidden layer growing mechanism. This behavior indicates that the feature extractor part of NADINE++ is able to generate a more generalizable feature representation \cite{wang2015transferring}. This situation can prevent sudden performance decrease whenever a concept change occurs. As a result, it triggers less addition of depth. Note that it is important to maintain the network complexity as low as possible to help for real-time deployment \cite{he2018amc}.
\begin{table}
\caption{Precision and Recall of NADINE++ and ResNet18 on Current-batch Prediction Scenario}
\label{result-currentbatch-precrec}
\begin{center}
\scalebox{1}{
\begin{sc}
\begin{tabular}{llccr}
\toprule
& Labels & Precision & Recall \\
\midrule
N++ & Normal & 0.92 $\pm$ 0.003 & 0.88  $\pm$ 0.016 \\
& Weaving & 0.84 $\pm$ 0.013 &\textbf{ 0.86 $\pm$ 0.000} \\
& Short-forming & \textbf{0.87 $\pm$ 0.000} & 0.90 $\pm$ 0.000 \\
\midrule
RN18 & Normal & 0.92 $\pm$ 0.011 & 0.88  $\pm$ 0.018 \\
& Weaving & 0.84 $\pm$ 0.013 & 0.83 $\pm$ 0.014$^\times$ \\
& Short-forming & 0.83 $\pm$ 0.011$^\times$ & 0.90 $\pm$ 0.013 \\
\bottomrule
\end{tabular}
\end{sc}}
\end{center}
\end{table}

Next, we present the precision and recall results of NADINE++ in Table \ref{result-currentbatch-precrec} to further evaluate its performance in predicting the mold quality. Overall, it is obvious that all classes achieved high recall with high precision. It was higher than 0.80 in all classes. Further, NADINE++ (N++) achieved statistically better F-score (P $<$ 0.05) than ResNet18 (RN18) in weaving and short-forming labels at 0.85 $\pm$ 0.006 and 0.88 $\pm$ 0.000 respectively, whereas ResNet18 was at 0.84 $\pm$ 0.012 and 0.86 $\pm$ 0.010. Note that F-score can be obtained easily from precision and recall. Intuitively, it measures how well NADINE++ predicts a class without confusion. This finding signifies that NADINE++'s prediction is not biased to one of the labels. This aspect is important in a prediction task \cite{virtanen2019precision}. Too many false negative and false positive diagnoses will reduce the production cost efficiency as the operator frequently needs to stop, to inspect and to adjust the machine.

\subsection{Ablation Study} Since NADINE++ combines three main components, it has something in common with existing methods in the literature. As a result, it is required to study the effect of removing or adding components to present additional insight into what makes NADINE++ better. Specifically, we quantify the effect of \textit{A}) reducing the number of 2-D CNN block layers from 4 into 2; \textit{B}) removing 1-D CNN, which makes NADINE++ unable to process sensor data; \textit{C}) turning off the evolving mechanism. The ablations were carried out in the second scenario; the results are depicted in Table \ref{result-ablation}. We found that every component contributes to NADINE++'s predictive performance. Another significant outcome from the same table is that reducing the number of 2-D CNN layer dramatically deteriorates NADINE++'s performance. This is understood as 2-D CNN plays an important role to extract useful features from image data which makes the classification task easier. Further, from an information-theoretic view for deep learning the nature of ResNet architecture, which incorporates residual connection, is able to prevent loss of information which makes deep network training convenient \cite{zhang2018information}.

\begin{table}
\caption{Classification performance on Ablation Study}
\label{result-ablation}
\begin{center}
\scalebox{1}{
\begin{sc}
\begin{tabular}{cccr}
\toprule
Ablation & Depth &  Acc. (\%) \\
\midrule
Original & 20.0 $\pm$ 0.47 & \textbf{87.1 $\pm$ 0.74}\\
\textit{A} & 14.0 $\pm$ 1.41 & 83.4 $\pm$ 1.26$^\times$ \\
\textit{B} & 20.7 $\pm$ 1.06 & 86.3 $\pm$ 1.70 \\
\textit{C} & 18 & 86.9 $\pm$ 0.57\\
\bottomrule
\end{tabular}
\end{sc}}
\end{center}
\centering{}\footnotesize{\textit{A}: Reduce the number of 2-D CNN block layer, \textit{B}: Without 1-D CNN, \textit{C}: Without evolving mechanism.}
\end{table}

\section{Conclusion}
We proposed NADINE++ a multi-modal deep learning approach for quality monitoring in streaming environments. NADINE++ is able to process both time-series and visual information utilizing 1-D CNN and 2-D CNN. The evolving MLP classifier was put forward to handle concept drift. The results indicate that NADINE++ achieved the best performance in all experiment scenarios. In terms of accuracy, it offered a 4.68\% improvement on average. We also found that the evolving mechanism succeeded to generate competitive network structures. In future work, we are interested in incorporating additional ideas from time-series prediction and continuing to explore which mechanisms result in effective methods. The code, raw numerical results, supplementary materials and the injection molding machine dataset can be accessed in this link \url{https://github.com/ContinualAL/NADINE-IJCNN2021}.

\section*{Acknowledgment}
This project is financially supported by NRF, Republic of Singapore under IAFPP in the AME domain (contract no.: A19C1A0018). The authors would like to thank Lee Wen Siong and Adithya Venkatadri Hulagadri for their assistance.




%

\bibliographystyle{IEEEtran}
\bibliography{mbibfile}

\end{document}


%
\title{Autonomous Deep Quality Monitoring \\in Streaming Environments}

\author{\IEEEauthorblockN{Andri Ashfahani}
\IEEEauthorblockA{School of Computer Science and\\Engineering\\
Nanyang Technological University\\
Singapore\\
Email: andriash001@e.ntu.edu.sg}
\and
\IEEEauthorblockN{Mahardhika Pratama}
\IEEEauthorblockA{School of Computer Science and\\Engineering\\
Nanyang Technological University\\
Singapore\\
Email: mpratama@ntu.edu.sg}
\and
\IEEEauthorblockN{Edwin Lughofer}
\IEEEauthorblockA{Department of Knowledge-Based\\ and Mathematical Systems\\ Johannes Kepler University\\ Austria}
\and
\IEEEauthorblockN{E. Y. K. Yee}
\IEEEauthorblockA{Singapore Institute of Manufacturing Technology\\Singapore}
}


%


\maketitle

\begin{abstract}
This document contains the supplementary material of our paper titled Autonomous Deep Quality Monitoring in Streaming Environments. It explains the injection molding machine which is involved in our experiment.
\end{abstract}


%
\IEEEpeerreviewmaketitle

\section{Injection Molding Machine}
\begin{figure}[!t]
\centerline{\includegraphics[scale=0.6]{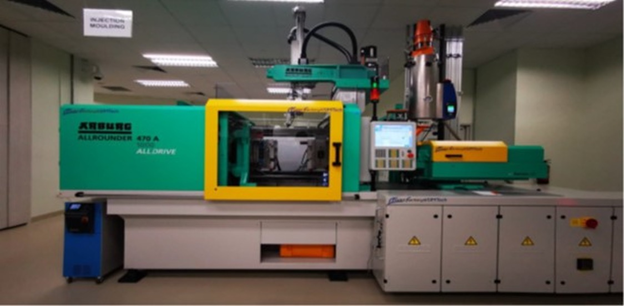}}
\caption{The injection molding machine used in this study.}
\label{fig:injmoldmachine}
\end{figure}

\IEEEPARstart{T}{he} injection molding machine is usually used to fabricate plastic products such as plastic trinkets, toys to automotive body parts, cell phone cases, water bottles, and containers \cite{kim2020zno}. The raw material, plastic, is injected through a nozzle into a mold cavity cooled and hardened as per the layout of the cavity \cite{bertschi1997injection}. Our experiment here makes use of the injection molding machine for the production of transparent mold as exhibited in Fig. \ref{fig:injmoldmachine}. This machine possesses many parameters where 48 of which are considered here and are listed in the appendix. Two important parameters, namely the holding pressure and the injection speed, are varied here to simulate concept drifts. That is, the holding pressure is varied to be 900, 700, 500, 300, 100 psi while the injection speed is set to be 60, 70, 80, 90, 100 rpm.

\begin{figure}
     \centering
     \begin{subfigure}
         \centering
         \includegraphics[scale=0.85]{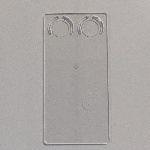}\\
         \footnotesize{(a)}\\
     \end{subfigure}
     \vspace{10pt}
     \begin{subfigure}
         \centering
         \includegraphics[scale=0.85]{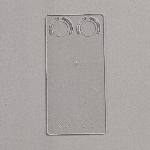}\\
         \footnotesize{(b)}\\
     \end{subfigure}
     \vspace{10pt}
     \begin{subfigure}
         \centering
         \includegraphics[scale=0.85]{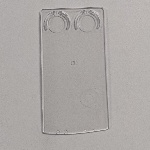}\\
         \footnotesize{(c)}
     \end{subfigure}
     \caption{The transparent mold quality produced by the injection molding machine: (a) Normal, (b) weaving and (c) short-forming. Weaving class indicates that the mold surface is uneven, whereas short-forming class specifies that the mold shape is not in a perfect rectangular shape.}
     \label{fig:material}
\end{figure}

Our online quality monitoring problem in this study considers two implementation scenarios: one-step-ahead and current-batch prediction problems. The former scenario requires a model to predict the next batch output $\hat{Y}_{k+1}$ based on the current batch sensor data $X_{k}$. It aims to estimate the possible next batch mold quality given the current batch machine condition from sensors. The latter problem focuses on predicting current batch mold quality $\hat{Y}_{k}$ based on the current batch sensor and image data. It functions as a monitoring tool where the operator is able to know the mold quality without turning-off the molding process. Both problems are multi-class classification problems consisting of 3 classes, i.e. good, weaving and short-forming. The number of data in good, weaving and short-forming classes is 1008, 1074 and 870, respectively. All those three mold qualities are displayed in Fig. \ref{fig:material}. It is RGB image with size 150 $\times$ 150 $\times$ 3. In practice, the injection molding machine requires around 9 minutes to complete a batch of molding process.

Table \ref{appendix-features} presents the list of input features captured from sensors. There are 48 input features from sensors utilized to predict the mold quality. To introduce concept changes, 2 variables were varied. Those are holding pressure and injection speed. Because of these variations, 3 classes indicating the mold quality were spotted by the operator. The observed qualities were normal, weaving and short-forming. All of those are presented in Fig. \ref{fig:material}. Finally, MinMax normalization was applied to have all features falling within the same range.
\begin{table}
\caption{The List of Features Captured from Sensors}
\label{appendix-features}
\begin{center}
\scalebox{0.9}{
\begin{sc}
\begin{tabular}{cl}
\toprule
Number & Description\\
\midrule
1	&	Screw volume, actual value	\\
2	&	Material cushion, actual value	\\
3	&	Dosage time, actual value	\\
4	&	Cycle time, actual value	\\
5	&	Mould heating circuit 1, actual value	\\
6	&	Mould heating circuit 2, actual value	\\
7	&	Mould heating circuit 3, actual value	\\
8	&	Mould heating circuit 4, actual value	\\
9	&	Mould heating circuit 5, actual value	\\
10	&	Mould heating circuit 6, actual value	\\
11	&	Plotting point 2 (Holding Pressure)$^*$	\\
12	&	Injection Speed$^*$	\\
13	&	Injection flow, actual value	\\
14	&	Switch-over volume, actual value	\\
15	&	Maximum injection pressure, actual value	\\
16	&	Injection time, actual value	\\
17	&	Cylinder heating zone 1, actual value	\\
18	&	Cylinder heating zone 2, actual value	\\
19	&	Cylinder heating zone 3, actual value	\\
20	&	Cylinder heating zone 4, actual value	\\
21	&	Cylinder heating zone 5, actual value	\\
22	&	Opening force, actual value	\\
23	&	Opening speed, actual value	\\
24	&	Oil temperature, actual value	\\
25	&	Mould temperature control unit 1, actual value	\\
26	&	Nozzle stroke, actual value	\\
27	&	Closing speed, actual value	\\
28	&	Advancement speed, actual value	\\
29	&	Retraction speed, actual value	\\
30	&	Mould protection force, actual value	\\
31	&	Temperature of support housing, actual value	\\
32	&	Circumferential speed, actual value	\\
33	&	Ejector pressure, nominal value	\\
34	&	Ejector pressure, actual value	\\
35	&	Nozzle 1 flow, nominal value	\\
36	&	Nozzle 1 pressure, actual value	\\
37	&	Injection torque, actual value	\\
38	&	Injection rotational speed, actual value	\\
39	&	Injection force of screw 1, actual value	\\
40	&	Dosage torque, actual value	\\
41	&	Dosage rotational speed, actual value	\\
42	&	Hydraulic accumulator pressure, actual value	\\
43	&	Charge pressure of accumulator, measured value	\\
44	&	Mould-entry time, actual value	\\
45	&	Part removal time, actual value	\\
46	&	Maximum injection pressure, actual value	\\
47	&	Back Pressure, Actual	\\
48	&	Clamping Force, Actual	\\
\bottomrule
\end{tabular}
\end{sc}}
\end{center}
\centering{}\footnotesize{$^*$: Indicate the varied features to introduce concept change.}
\end{table}

\bibliographystyle{IEEEtran}
\bibliography{mbibfile}
